\documentclass{article}

\PassOptionsToPackage{numbers,square,comma,sort}{natbib}
\usepackage{footnote}

\usepackage[preprint]{neurips_2022}

\usepackage[utf8]{inputenc} 
\usepackage[T1]{fontenc}    
\usepackage{hyperref}       
\usepackage{url}            
\usepackage{booktabs}       
\usepackage{amsfonts}       
\usepackage{nicefrac}       
\usepackage{microtype}      
\usepackage{xcolor}         
\usepackage{amssymb}

\usepackage{url}            
\usepackage{booktabs}       
\usepackage{amsmath}
\usepackage{amsthm}
\usepackage{graphicx}
\usepackage{diagbox}
\usepackage{tabularx}
\usepackage{bm}
\usepackage{bbm}
\usepackage{algorithm}
\usepackage{algorithmic}
\usepackage{setspace}
\usepackage{subfigure}

\usepackage{amsfonts}
\usepackage{float}
\usepackage{hyperref}
\usepackage{enumitem,graphicx}

\usepackage{multirow}

\makeatletter
\DeclareRobustCommand\onedot{\futurelet\@let@token\@onedot}
\def\@onedot{\ifx\@let@token.\else.\null\fi\xspace}

\usepackage{neurips_2022}

\title{Enhanced Few-Shot Class-Incremental Learning via Ensemble Models}

\newcommand*{\email}[1]{%
    \normalsize\href{mailto:#1}{#1}\par
    }

\author{
Mingli Zhu\textsuperscript{1}\thanks{This work was done when Mingli Zhu was interned at Tencent.} ,\ \ \ \ 
Zihao Zhu\textsuperscript{1}, \ \ \ \ 
Sihong Chen\textsuperscript{2}, \ \ \ \ 
Chen Chen\textsuperscript{2}, \ \ \ \
Baoyuan Wu\textsuperscript{1}\thanks{Corresponds to Baoyuan Wu (\email{wubaoyuan@cuhk.edu.cn}).} \\
\textsuperscript{1}School of Data Science, \\
The Chinese University of Hong Kong, Shenzhen (CUHK-Shenzhen), China \\
\textsuperscript{2}
Tencent TEG AI, Shenzhen, China
}

\begin{document}

\maketitle

\begin{abstract}
Few-shot class-incremental learning (FSCIL) aims to continually fit new classes with limited training data, while maintaining the performance of previously learned classes. The main challenges are overfitting the rare new training samples and forgetting old classes. While catastrophic forgetting has been extensively studied, the overfitting problem has attracted less attention in FSCIL. To tackle overfitting challenge, we design a new ensemble model framework cooperated with data augmentation to boost generalization. In this way, the enhanced model works as a library storing abundant features to guarantee fast adaptation to downstream tasks. Specifically, the multi-input multi-output ensemble structure is applied with a spatial-aware data augmentation strategy, aiming at diversifying the feature extractor and alleviating overfitting in incremental sessions. Moreover, self-supervised learning is also integrated to further improve the model generalization. Comprehensive experimental results show that the proposed method can indeed mitigate the overfitting problem in FSCIL, and outperform the state-of-the-art methods.

\end{abstract}

\section{Introduction\label{sec1}}

Traditional classification problems in machine learning assume that training data and test classes are fixed \cite{tang2004video,liu2006spatio}. 
However, this stringent assumption is often not fulfilled in many real-world situations \cite{hadsell2020embracing,pan2009survey}. In practice, the model trained on large-scale dataset needs to be able to quickly adapt to new classes while maintaining the performance of  previously learned classes \cite{parisi2019continual, li2017learning, kemker2017fearnet}. 
A more severe situation is that we can only use limited images to train the incremental model when collecting data is expensive or even impossible.
This is formally known as the Few-Shot Class-Incremental Learning (FSCIL) problem, which has attracted attention in recent years \cite{ayub2020cognitively, tao2020few, zhao2021mgsvf}.

Overfitting and catastrophic forgetting are two main challenges for FSCIL tasks. 
For example, suppose we have a model trained on base classes in the first session. When the new session comes, the pre-trained model needs to finetune on the new few-shot images. The new training on the few-shot instances will overwrite the decisive parameters and cause the model to be biased towards new classes \cite{ramasesh2020anatomy}. The overfitting problem happens when the deep network quickly remembers the limited samples without access to the distribution of the classes. Moreover, the \textbf{overfitting} problem will in turn exacerbate \textbf{catastrophic forgetting}. Firstly, novel classes neither have enough training samples nor enough contrastive samples to train with, thus the model will fast converge to the training samples with poor generalization. We analyze this phenomenon in our experimental section. The model cannot grasp the representative features. Thus, the performance of the fragile new learned classes will drop sharply as the learning process goes on, which behaves as severe catastrophic forgetting. Secondly, in few-shot learning sessions, new samples are probably dispersed, lying between representations of previous classes and occupying the feature space of old classes \cite{caccia2022new} instead of learning their own representative features. Then the model tends to confuse the old and new classes. The loss of discriminability of the model aggravates catastrophic forgetting.
To sum up, overfitting is a stronger obstacle than catastrophic forgetting in FSCIL, and addressing overfitting can help facilitate a better incremental learning paradigm.

Existing studies \cite{rebuffi2017icarl, hou2019learning,tao2020few, chen2020incremental, zhao2021mgsvf, yang2021learnable, mazumder2021few, akyurek2021subspace, shi2021overcoming} mainly focus on tackling catastrophic forgetting, while ignoring the overfitting problem fundamentally. 
 Replay-based \cite{rebuffi2017icarl, hou2019learning,dong2021few,shi2021overcoming} methods borrow ideas from continual learning that generally save some old exemplars and retrain them with new samples in incremental sessions. 
Regularization-based methods \cite{tao2020few, chen2020incremental,  mazumder2021few, akyurek2021subspace} impose various constraints straight on model parameters or use knowledge distillation to defy forgetting. 
 Recently there are some relation-based works that decouple the training process and model the relation of these prototypes to suitably insert new classes \cite{dong2021few, zhu2021self,  ye2021learning, zhang2021few}. 
 However, all these works mainly focus on preventing catastrophic forgetting while never prioritizing another underlying problem --- overfitting. To this end, how to avoid overfitting and in turn overcome forgetting is the focus of our work.

In this work, we propose a lightweight ensemble-based framework integrated with well-designed data augmentation and self-supervision strategies to address the above challenges.
A single model is relatively weak to provide adequate knowledge for incremental sessions. 
Inspired by model ensemble which has shown great strength of  improving generalization and solving overfitting in traditional machine learning scenarios \cite{dietterich2000ensemble, chowdhury2021few}, we apply ensemble model as our architecture to extract more knowledge from data.
However, introducing complex ensemble models will introduce too many parameters which in turn leads to more severe overfitting.
Thus we adopt a \textbf{lightweight, multi-input multi-output ensemble framework} with a shared core network as our backbone. 
In this way, the backbone can work as an abundant feature library that accommodates the downstream tasks. This well-trained backbone can provide diverse distinguished feature templates in advance, so there is no need to make extensive modifications to the model to overfit new instances.

Furthermore, we propose data augmentation and self-supervision strategies to enhance our ensemble framework. 
On the one hand, although our backbone can identify diverse features, it cannot distinguish the object from the images because of the lack of abundant negative samples in few-shot sessions. To tackle this, we propose a \textbf{spatial-aware data augmentation} strategy. Different from existing mixed sample data augmentation (MSDA) method, it is a background-level augmentation that introduces diversity and protects the principal part of samples simultaneously.
On the other hand, to make the model concentrate on generic and universal representations and promote generalization, we integrate \textbf{mix-feature compatible self-supervised learning} with our proposed ensemble framework. With the help of self-supervised learning, the overfitting problem will be mitigated in the incremental sessions.
Our main contributions can be summarized as follows:
\begin{itemize}
\item We propose a lightweight multi-input multi-output ensemble framework to solve overfitting for FSCIL task. To the Best of our knowledge, we are the first to utilize model ensembling for FSCIL.
\item We propose a novel spatial-aware MSDA method to prevent overfitting on new rare samples.
\item We propose mix-feature compatible self-supervised learning to enhance the performance of representation learning and better accommodate downstream tasks.
\item Exhaust experiments on the CIFAR100, CUB200 and \textit{mini}ImageNet datasets demonstrate that our method can outperform the state-of-the-art method with remarkable promotion.
\end{itemize}

\section{Related work}

\paragraph{Few-shot learning and incremental learning.} Few-shot learning \cite{wang2020generalizing}  considers that we have access to limited data and a pre-trained model. The model has to be very adaptable to the small number of novel data. Generally, there are mainly two categories: metric-based methods and optimization-based methods. Metric-based approaches first pre-train a backbone as the feature extractor and utilize some distance metric functions over the embeddings to measure similarities between samples \cite{snell2017prototypical, sung2018learning, vinyals2016matching, zhang2020deepemd}. Optimization-based methods aim to enable the model to fast adapt to new classes with limited data \cite{finn2017model, jamal2019task, chen2020new}. Most of these methods concentrate on fast learning of new classes while ignoring the recognition of initial classes. More recently, some works have attempted to research on generalized few-shot learning, where the performance of base and novel classes are considered \cite{mazumder2021few, kukleva2021generalized}. Incremental learning has been extensively studied for many years since it is still challenging for neural networks to learn tasks in sequence\cite{parisi2019continual, delange2021continual, li2017learning, masana2020class, hou2019learning, rebuffi2017icarl, lopez2017gradient}. It can be divided into task-incremental learning \cite{li2017learning, hu2018overcoming} and class-incremental learning. Compared to the former, class-incremental learning is more challenging because a unified classifier is used for all classes without task information. This is a thriving field where many different methods have been proposed in recent years. Replay-based methods use knowledge distillation to transfer old knowledge, where old exemplars are saved in a memory buffer \cite{hou2019learning, rebuffi2017icarl, lopez2017gradient} or a generator will be trained for further use \cite{shin2017continual, lavda2018continual}. Regularization-based methods introduce extra regularization terms in the loss function \cite{ chaudhry2018riemannian, kirkpatrick2017overcoming, li2017learning, castro2018end}. For example, EWC \cite{kirkpatrick2017overcoming} imposes strict constraints on these important parameters while MgSvF \cite{zhao2021mgsvf} decomposes parameters from frequency's perspective.

\paragraph{Few-shot class-incremental learning.}   FSCIL gradually raises attention in recent several years \cite{ayub2020cognitively, tao2020few, zhu2021self, dong2021few, mazumder2021few, zhou2022forward, kukleva2021generalized, chen2020incremental, akyurek2021subspace, shi2021overcoming}. It aims to learn continually with limited training data in incremental sessions. TOPIC \cite{tao2020few} initiatively uses neural gas to preserve the topology structure of learned knowledge. FSLL \cite{mazumder2021few} selects important parameters with high absolute magnitude and keeps them close to the original value to preserve old knowledge. CEC \cite{zhang2021few} fixes the embedding network and proposes a continually evolved classifier based on graph attention network to adapt to new classes. F2M \cite{shi2021overcoming} overcomes catastrophic forgetting by searching for flat local minima and constraints the region of parameters' change. FACT \cite{zhou2022forward} considers forward compatible training, which preserves embedding space for new classes in the pre-training stage.

\paragraph{Data augmentation.}
Data augmentation (DA) aims to reduce overﬁtting by creating artificial samples. It has been broadly used in long-tailed classification \cite{liu2020deep, chu2020feature} and few-shot learning scenarios \cite{schwartz2018delta, zhang2018metagan, yang2021free, wang2018low}. Traditional DA methods impose different transformations to produce more data, \textit{e.g.}, flipping, cropping, or rotation. Recently, mixed sample data augmentation (MSDA) methods have been studied from which diverse samples are generated \cite{devries2017improved, zhang2017mixup, yun2019cutmix}. In \cite{chu2020feature}, novel samples are generated by fusing class-generic features from head classes and class-specific features from tailed data using class activation maps. In \cite{wang2018low}, they synthesize new data by training a generative model to alleviate data insufﬁciency.

\section{Methodology}
\begin{figure}
\centering
\includegraphics[width=1\textwidth]{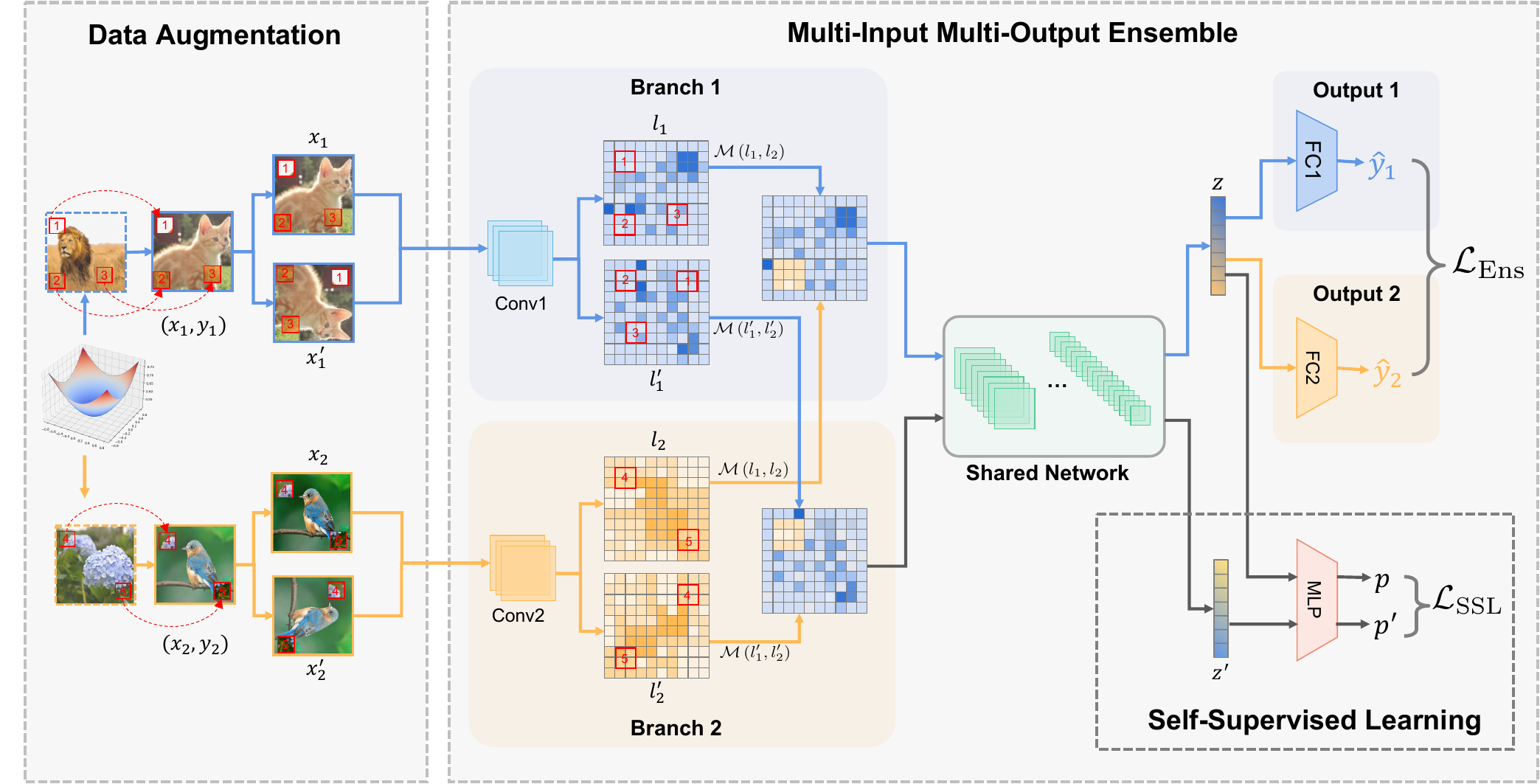}
\caption{The Illustration of our method. It consists of three components: (1) Ensemble model structure. The multi-input multi-output network is used as the backbone. (2) Spatial-aware data augmentation. The background of images is augmented to alleviate overfitting problem for few-shot learning; (3) Self-supervision loss. A contrastive learning loss is used to boost the performance of our backbone.}
\label{frame}
\end{figure}

In this section, we introduce our method for few-shot class-incremental learning. We first give a formal problem statement in Section \ref{sec3.1}. Then we introduce the whole framework in Section \ref{sec3.2}. Figure \ref{frame} shows the overall training pipeline. It consists of three components: (1) Ensemble model structure. We use this model as our backbone to provide diverse features templates in advance; (2) Spatial-aware data augmentation. This well-designed augmentation method can diversify the background of images and alleviate overfitting problem for few-shot learning; (3) Self-supervision loss. We integrate SSL with model ensembling in our architecture to boost the performance of our backbone.

\subsection{Problem formulation\label{sec3.1}}
We define the FSCIL problem as follows. The data are coming in a stream of labelled datasets $D^{(0)},D^{(1)},D^{(2)},...,D^{(T)}$, where $D^{(t)}$ is composed of a support set $S^{t}$ and a query set $Q^{t}$. $C^{(t)}$ denotes the set of classes in each training session and $C^{(i)}\cap C^{(j)}= \emptyset$ for any $i\ne j$, which means the coming classes are always different from the former ones. When $t=0$, the training set is a large-scale dataset. A model trained on $S^{0}$ is named a base model. Then the same model is trained on the coming new classes with a unified classifier, where the coming dataset $S^{t}$ consists of a limited amount of data and N-way K-shot is a commonly-used setting. In another word, it contains $N$ classes and $K$ samples per class. After each session, the model is tested on the whole test dataset $Q^{(\leq t)}=\cup_{j=1}^{t} Q^{(j)}$. The goal of FSCIL is to finetune the same model with the novel 
scarce training data and the model can perform well on the whole classes without forgetting. The challenge of this problem is to prevent forgetting old classes and alleviate overfitting on new few-shot classes.

\subsection{Overall framework\label{sec3.2}}
Our goal is to devise a backbone that can work as a feature library storing abundant features for the incremental learning sessions. Motivated by the power of ensemble model to boost generalization, we use ensemble model as our main structure. The whole training pipeline is shown in Figure \ref{frame}. As introduced in Section \ref{sec1}, the aim is to train a backbone that can provide diverse distinguished features template so as to accommodate the learning of new classes. However, things are more troublesome in FSCIL scenario. Since too few samples are provided in each incremental session, overmuch parameters will 
aggravate overfitting. To balance the two concerns, we adopt the multi-input multi-output structure as our backbone inspired by recent works \cite{rame2021mixmo,havasi2020training}. 

Furthermore, we integrate our method with self-supervision to further alleviate the overfitting problem. Self-supervised learning forces the model to concentrate on generic and universal representations. However, the multi-input multi-output structure contradicts the training paradigm in contrastive learning that needs to produce different views of one image. Specifically, since feature fusion is implemented, it is hard to keep two views of one image in the training process. To this end, we creatively develop a new SSL strategy that fits well with model ensembling. As shown in Figure \ref{frame}, the mixing masks do the same transformation as the images do. This measure can keep the relative position of each branch and the two mixed features can be seen as generated from two transformations of one image. Moreover, although our backbone can identify diverse features, it cannot distinguish the object from the images because of the lack of abundant negative samples. Thus we design a spatial-aware data augmentation module added to the input level in incremental learning sessions. In particular, we design a new patchmix data augmentation strategy that can diversify the background. In this way, the distribution of few-shot data can be calibrated.

\subsection{Our method\label{sec3.3}}
\subsubsection{Multi-in multi-output ensemble model \label{sec3.3.1}}
We use ensemble model as our backbone to improve the generalization. Instead of combining a library of pre-train models, we use a lightweight ensemble model inspired by recent works\cite{rame2021mixmo,havasi2020training}. We describe our framework followed by describing the core components here.

We adopt a multi-input multi-output ensembling framework to solve image classification problem. Two branches of subnetwork are used here and in the experiments for convenience. As shown in Figure \ref{frame}, a core backbone is shared by the two subnetworks. Each branch has its own convolutional layer in the first layer and fully-connected layer in the last. Thus model ensembling is realized almost for free. In \textbf{inference} stage, two identical images are sent to the two branches and the prediction result is the combination of the two classifiers.

For the feature mixing mechanism, we adopt the same strategy like CutMix as in \cite{rame2021mixmo}. Cutmix \cite{yun2019cutmix} realizes data augmentation by pasting one square patch from one image onto another in image level. Similar to CutMix, we combine the two features extracted by the two subnetworks in the feature level using a binary mask. Then the mixed feature is encoded through the shared network. After that, the two fully-connected layers are responsible for each input and predict different labels. The training loss is the sum of two cross-entropy loss weighted by a parametrized function. As shown in Figure \ref{frame}, let ${(x_{1},y_{1}),(x_{2},y_{2})}$ be two augmented training samples. Firstly they are separately encoded into the shared latent space with each convolutional layer $c_{1}$ and $c_{2}$. Then the mixing block $\mathcal{M}$ produces a binary mask $\mathbbm{1}_{\mathcal{M}}$ with area ratio $\kappa \sim Beta(\alpha,\alpha)$ to combine the two features $l_{1}=c_{1}(x_{1})$, $l_{2}=c_{2}(x_{2})$ as follows:
\begin{equation}
\mathcal{M}\left(l_{1}, l_{2}\right)=2\left[\mathbbm{1}_{\mathcal{M}} \odot l_{1}+\left(\mathbbm{1}-\mathbbm{1}_{\mathcal{M}}\right) \odot l_{2}\right].
\end{equation}
After that, the core network $\mathcal{C}$ encodes the mixed feature, followed by two fully-connected layer $d_{1}$ and $d_{2}$. $d_{1}$ predicts $\hat{y}_{1}= d_{1}[\mathcal{C}(\mathcal{M}\left(l_{1}, l_{2}\right))]$ and $d_{2}$ predicts $\hat{y}_{2}$. The network is trained with the following loss function:
\begin{equation}
\mathcal{L}_{\mathrm{Ens}}=w_{r}(\kappa) \mathcal{L}_{\mathrm{CE}}\left(y_{1}, \hat{y}_{1}\right)+w_{r}(1-\kappa) \mathcal{L}_{\mathrm{CE}}\left(y_{2}, \hat{y}_{2}\right).
\end{equation}

We keep the same parametrized function as in \cite{rame2021mixmo}:
\begin{equation}
w_{r}(\kappa)=2 \frac{\kappa^{1 / r}}{\kappa^{1 / r}+(1-\kappa)^{1 / r}}. \label{eq2}
\end{equation}

For the cross-entropy loss, we use cosine similarity to reduce bias. This is the whole scheme of our framework. In incremental learning sessions, we keep the same training strategy.

\subsubsection{Integration with spatial-aware data augmentation\label{sec3.3.2}}
In order to bring diversity and calibrate the few-shot classes that the former strategy has not considered, we impose data augmentation module on the input layer. Different from traditional MSDA, our main idea is to enrich the background of the limited novel data so as to calibrate the distribution of few-shot classes. A natural approach is to randomly replace some patches of the image with other images. Considering the limited number of samples, we want to keep the object of samples intact. We assume that the background of one image appears in the side part of the whole image. Thus we propose a special sampling strategy. Assume image $B$ is used to augment image $A$. Firstly image $B$ is divided into $N^{2}$ grid patches. These patches are sampled by a bowl-like distribution, which is specifically a normalized negative log of binary Gaussian distribution. The number of patches to select is also random. Then the sampled binary mask is imposed on the two images and a new background-augmented image is achieved. In our method, some samples from the base session are saved to augment the new classes.

\subsubsection{Integration with self-supervision\label{sec3.3.3}}
To boost the performance of our backbone and force the model to learn more features instead of overfitting on the base training data, we use self-supervision in the base session. Considering contrastive learning methods suffer from the need of a lot of negative examples, we propose a mix-feature compatible approach based on a recent work \cite{bardes2021vicreg} that doesn't require a large budget of batches. The contrastive loss is used as an auxiliary task to our problem. While we use the feature mixing block in the ensembling model, common contrastive method cannot be used directly in our framework. As shown in Figure \ref{frame}, we modify the SSL method in this way: for the two different images $x_{1}$ and $x_{2}$, we sample two transformations $t$ and $t^{\prime}$ from a pre-defined transformation distribution $\mathcal{T}$. Then the transformations are imposed on the two images and four images are generated. We assume that the shallow layer can maintain the space location compared to the original image. When the mixing block randomly produces a binary mask $\mathbbm{1}_{\mathcal{M}}$, we impose the same two transformations to the mask respectively. In this way, we maintain the consistency of two different views of mixed features. Then the two mixed features are sent into the core network to produce two representation $z$ and $z^{\prime}$. Finally, they are sent into a feature projector to compute the SSL loss.

The whole objective function is given by:
\begin{equation}
	\mathcal{L} = \mathcal{L}_{\mathrm{Ens}} + \gamma \mathcal{L}_{\mathrm{SSL}},
\end{equation}
where $\gamma$ is the hyper-parameter controlling the importance of the two terms.

\begin{table}[!h]
\caption{Comparison with the state-of-the-art on \textit{mini}ImageNet dataset.}
\vskip -0.2in
\label{mini_sota}
\begin{center}
\resizebox{\textwidth}{!}{
\begin{tabular}{lccccccccccc}
\toprule
\multirow{2}{*}{Method} & \multicolumn{9}{c}{Acc. in each session(\%). \textsuperscript{$\ast$} indicates our re-implementation} & \multirow{2}{*}{\shortstack{Our Acc.\\ improvement}} & \multirow{2}{*}{PD$\downarrow$}\\
\cmidrule(r){2-10}
 & 1 & 2 & 3 & 4 & 5 & 6 & 7 & 8 & 9 &  &  \\
 \midrule
Finetune	&	61.31 	&	27.22 	&	16.37 	&	6.08 	&	2.54 	&	1.56 	&	1.93 	&	2.60 	&	1.40 	&	\textbf{+55.13} & 59.91 	 	\\
iCaRL\cite{rebuffi2017icarl}	&	61.31 	&	46.32 	&	42.94 	&	37.63 	&	30.49 	&	24.00 	&	20.89 	&	18.80 	&	17.21 	&	\textbf{+39.32}  & 44.10 	 	\\
Rebalance\cite{hou2019learning} & 72.30 & 66.37 & 61.00 & 56.93 & 53.31 & 49.93 & 46.47 & 44.13 & 42.19 & \textbf{+14.34} & 30.11 \\
TOPIC\cite{tao2020few}	&	61.31 	&	50.09 	&	45.17 	&	41.16 	&	37.48 	&	35.52 	&	32.19 	&	29.46 	&	24.42 	&	\textbf{+32.11} & 36.89 		\\
SPPR\cite{zhu2021self}	&	61.45 	&	63.80 	&	59.53 	&	55.53 	&	52.50 	&	49.60 	&	46.69 	&	43.79 	&	41.92 	&	\textbf{+14.61} & 19.53  	\\
EKDIL\cite{dong2021few} & 62.90 & 58.21 & 54.95 & 52.47 & 48.35 & 47.18 & 44.26 & 43.16 & 40.79 & \textbf{+15.74} & 22.11 \\
CEC\cite{zhang2021few}	&	72.00 	&	66.83 	&	62.97 	&	59.43 	&	56.70 	&	53.73 	&	51.19 	&	49.24 	&	47.63 	&	\textbf{+8.90}  & 24.37 		\\
FSLL\cite{mazumder2021few}	&	66.48 	&	61.75 	&	58.16 	&	54.16 	&	51.10 	&	48.53 	&	46.54 	&	44.20 	&	42.28 	& \textbf{+14.25} &	24.20 	\\
F2M\cite{shi2021overcoming}	&	72.05 	&	67.47 	&	63.16 	&	59.70 	&	56.71 	&	53.77 	&	51.11 	&	49.21 	&	47.84 	&	\textbf{+8.69} & 24.21	 	\\
FACT\cite{zhou2022forward}\textsuperscript{$\ast$} & 75.63 & 70.49 & 66.37 & 62.61 & 59.10 & 56.00 & 53.04 & 51.04 & 49.29 & \textbf{+7.24} & 26.34  \\
\midrule
Ours	&	\textbf{81.28} 	&	\textbf{74.29} & \textbf{70.07} & \textbf{66.51} & \textbf{63.80} & \textbf{61.40} & \textbf{57.99} & \textbf{57.04} & \textbf{56.53} & - &	24.75\\

\bottomrule
\end{tabular}
}
\end{center}
\end{table}

\section{Experiments}

In this section, we implement our proposed method for incremental few-shot learning on three benchmark datasets and demonstrate its effectiveness by comparison with the state-of-the-art methods. We first describe the details of the datasets and experimental setup. Then we compare our method with state-of-the-art methods on the benchmarks. Finally, we conduct comprehensive experiments to verify the effectiveness of individual components of our method and study their properties.

\subsection{Experimental setup\label{sec4.1}}
\textbf{Datasets:} We evaluate our method on CIFAR100 \cite{krizhevsky2009learning}, \textit{mini}ImageNet \cite{russakovsky2015imagenet} and CUB200-2011 \cite{wah2011caltech} following the benchmarks \cite{tao2020few}. For details, CIFAR100 contains 60,000 images from 100 classes, with 500 images per class for training and 100 images per class for testing. Each image has a size of $32 \times 32$. \textit{mini}ImageNet is a subset of ImageNet, which contains 60,000 images over 100 classes, each class with 500 images for training and 100 images for testing. Each image has a size of $84 \times 84$. CUB200 is a fine-grained dataset in 200 classes, which contains 11788 for training and testing. Each image has a size of $224 \times 224$. For CIFAR100 and \textit{mini}ImageNet, 60 classes are selected as base classes and the remaining 40 classes are new classes. The new classes are constructed as eight 5-way 5-shot incremental learning tasks. For CUB200, 200 classes are divided into 100 base classes and 100 new classes. The new classes are constructed as ten 10-way 5-shot incremental learning tasks. The test datasets just remain to be the original ones. We use the same training split as in \cite{tao2020few} (including the separation of classes and the few-shot training samples) for the three datasets for a fair comparison. 

\textbf{Comparison details.} We compare our method with existing FSCIL methods on the three benchmarks. We compare to classical class-incremental method iCaRL \cite{rebuffi2017icarl}, Rebalance \cite{hou2019learning} and state-of-the-art FSCIL methods TOPIC \cite{tao2020few}, ERDIL \cite{dong2021few}, SPPR \cite{zhu2021self}, FSLL \cite{mazumder2021few}, CEC \cite{zhang2021few}, F2M \cite{cheraghian2021semantic} and FACT\cite{zhou2022forward}. We also show the results by finetuning the model as `finetune'. 

\textbf{Training details.} 
We adopt ResNet18 \cite{he2016deep} as our backbone for all the three datasets as in \cite{tao2020few}. All the experiments are conducted with PyTorch. We train the base model with batch size 128 on CIFAR100 and \textit{mini}ImageNet and 64 for CUB200 for 400 epochs. The model is trained using SGD with momentum and weight decay. For the ensemble model, we use two branches as in \cite{rame2021mixmo}. For PatchMix, the images are divided into $8\times 8$ grids and $3\sim 5$ patches are selected uniformly with a probability distribution described in Section \ref{sec3.3.1} for CIFAR100 and \textit{mini}ImageNet. Images are divided into $14\times 14$ grids for CUB200. For SSL method, we set the hyper-parameter $\gamma$ to $0.2$ for CIFAR100 and \textit{mini}ImageNet datasets and $0.1$ for CUB200. A model pre-trained on Imagenet is used for CUB200. We only use SSL in the base session because we find that SSL loss will induce instability in few-shot learning sessions. PatchMix is used for all the sessions. More details of the experimental setting can be found in Appendix. To select typical exemplars, we conduct K-means on the output of the feature extractor and select 5 images for each class. These exemplars will participate in the following training process. We also initialize the new parameters of the classifiers by randomly selecting two training data from each new class and extracting the features.

\begin{table}[!t]
\caption{Comparison with the state-of-the-art on CIFAR100 dataset.}
\vskip -0.2in
\label{cifar_sota}
\begin{center}
\resizebox{\textwidth}{!}{
\begin{tabular}{lccccccccccc}
\toprule
\multirow{2}{*}{Method} & \multicolumn{9}{c}{Acc. in each session(\%). \textsuperscript{$\ast$} indicates our re-implementation} & \multirow{2}{*}{\shortstack{Our Acc.\\ improvement}} & \multirow{2}{*}{PD$\downarrow$}\\
\cmidrule(r){2-10}
 & 1 & 2 & 3 & 4 & 5 & 6 & 7 & 8 & 9 &  &  \\
 \midrule
Finetune	&	64.00 	&	37.30 	&	16.17 	&	10.05 	&	8.24 	&	4.56 	&	4.24 	&	4.09 	&	3.29 	&		\textbf{+47.6} & 60.71 	\\
iCaRL\cite{rebuffi2017icarl}	&	64.10 	&	53.28 	&	41.69 	&	34.13 	&	27.93 	&	25.06 	&	20.41 	&	15.48 	&	13.73 	&	\textbf{+37.16}	& 50.37 \\
Rebalance\cite{hou2019learning}  & 74.45 & 67.74	& 62.72	& 57.14	& 52.78	& 48.62	& 45.56	& 42.43	& 39.22 &  \textbf{+11.67} & 35.23 \\

TOPIC\cite{tao2020few}	&	64.10 	&	56.03 	&	47.89 	&	42.99 	&	38.02 	&	34.60 	&	31.67 	&	28.35 	&	25.86 	&	\textbf{+25.03}	& 38.24 \\
SPPR\cite{zhu2021self}	&	63.97 	&	65.86 	&	61.31 	&	57.60 	&	53.39 	&	50.93 	&	48.27 	&	45.36 	&	43.32 	&	\textbf{+7.57} & 20.65 	\\
EKDIL\cite{dong2021few} & 73.50 & 68.20 & 66.49 & 63.67 & 60.08 & 57.25 & 54.61 & 52.86 & 50.04 & \textbf{+0.85} & 23.46 \\
CEC\cite{zhang2021few}\textsuperscript{$\ast$}	&	66.40 	&	62.32 	&	58.37 	&	55.04 	&	52.11 	&	49.46 	&	47.21 	&	45.10 	&	43.30 	&	\textbf{+7.59} & 23.10 	\\
FSLL\cite{mazumder2021few}	&	64.10 	&	55.85 	&	51.71 	&	48.59 	&	45.34 	&	43.25 	&	41.52 	&	39.81 	&	38.16 	&	\textbf{+12.73}	& 25.94 \\
F2M\cite{shi2021overcoming}	&	71.45 	&	68.10 	&	64.43 	&	60.80 	&	57.76 	&	55.26 	&	53.53 	&	51.57 	&	49.35 	&	\textbf{+1.54} & 	22.10	\\
FACT\cite{zhou2022forward}\textsuperscript{$\ast$} &	69.90 	&	65.26 	&	60.94 	&	57.55 	&	54.10 	&	51.28 	&	48.80 	&	46.62 	&	44.46 		&	\textbf{+6.43} &	25.44 \\
 \midrule
Ours	&	\textbf{76.60} &	\textbf{71.57}	&	\textbf{66.89}	&	\textbf{62.63}	&	\textbf{60.22}	&	\textbf{57.48}	&	\textbf{55.22}	&	\textbf{53.16}	&	\textbf{50.89}	& -	&	25.71	\\
\bottomrule
\end{tabular}
}
\end{center}
\end{table}

\begin{table}[!t]
\caption{Comparison with the state-of-the-art on CUB200 dataset.}
\vskip -0.2in
\label{cub_sota}
\begin{center}
\resizebox{\textwidth}{!}{
\begin{tabular}{lccccccccccccc}
\toprule
\multirow{2}{*}{Method} & \multicolumn{11}{c}{Acc. in each session(\%). \textsuperscript{$\ast$} indicates our re-implementation} & \multirow{2}{*}{\shortstack{Our Acc.\\ improvement}} & \multirow{2}{*}{PD$\downarrow$}\\
\cmidrule(r){2-12}
 & 1 & 2 & 3 & 4 & 5 & 6 & 7 & 8 & 9 & 10 & 11 &  &  \\
  \midrule
Finetune	&	68.68	&	43.7	&	25.05	&	17.72	&	18.08	&	16.95	&	15.10	&	10.06	&	8.93	&	8.93	&	8.47 & \textbf{+48.62} & 60.21\\
iCaRL\cite{rebuffi2017icarl}	&	68.68	&	52.65	&	48.61	&	44.16	&	36.62	&	29.52	&	27.83	&	26.26	&	24.01	&	23.89	&	21.16 &	\textbf{+35.93}& 47.52\\
Rebalance\cite{hou2019learning} & 77.44 & 58.10 & 50.15 & 44.80 & 39.12 & 34.44 & 31.73 & 29.75 & 27.56 & 26.93 & 25.30 & \textbf{+31.79} & 52.14 \\
TOPIC\cite{tao2020few}	&	68.68	&	62.49	&	54.81	&	49.99	&	45.25	&	41.4	&	38.35	&	35.36	&	32.22	&	28.31	&	26.28 & \textbf{+30.81} & 42.40  \\
SPPR\cite{zhu2021self}	&	68.68	&	61.85	&	57.43	&	52.68	&	50.19	&	46.88	&	44.65	&	43.07	&	40.17	&	39.63	&	37.33 & \textbf{+19.76}	& 27.75 \\
EKDIL\cite{dong2021few} & 73.52 & 71.09 & 66.13 & 63.25 & 59.49 & 59.89 & 58.64 & 57.72 & 56.15 & 54.75 & 52.28 & \textbf{+4.81} & 21.24 \\
CEC\cite{zhang2021few}	&	75.85	&	71.94	&	68.5	&	63.5	&	62.43	&	58.27	&	57.73	&	55.81	&	54.83	&	53.52	&	52.28 & \textbf{+4.81} & 23.57 	\\
FSLL\cite{mazumder2021few}	&	72.77 	&	69.33 	&	65.51 	&	62.66 	&	61.10 	&	58.65 	&	57.78 	&	57.26 	&	55.59 	&	55.39 	&	54.21 & \textbf{+2.88} & 18.56	\\
F2M\cite{shi2021overcoming}	&	77.13 	&	73.92 	&	70.27 	&	66.37 	&	64.34 	&	61.69 	&	60.52 & 59.38 	&	57.15 	& 56.94	& 55.89 & \textbf{+1.20} & 21.24 \\
FACT\cite{zhou2022forward}\textsuperscript{$\ast$} & 77.26 & 73.16 & 69.66 & 65.38 & 64.81 & 61.32 & 60.75 & 59.651 & 57.28 & 57.22 & 56.09 & \textbf{+1.00} & 21.17  \\
\midrule
Ours	&	\textbf{80.83} & \textbf{76.77} & \textbf{71.98} & \textbf{68.27} & \textbf{67.78} & \textbf{66.27} & \textbf{63.42} & \textbf{61.98} & \textbf{60.05} & \textbf{58.65} & \textbf{57.09} & - & 23.74 \\
\bottomrule
\end{tabular}
}
\end{center}
\end{table}

\begin{figure}
\centering
\includegraphics[width=\textwidth]{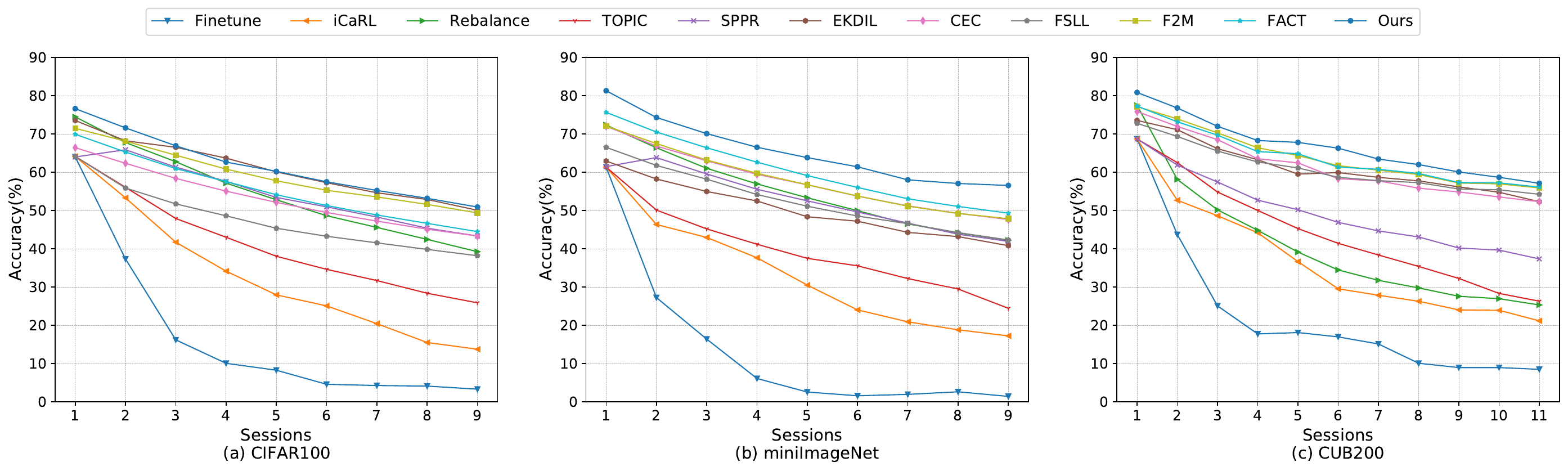}
\caption{Comparison of the proposed method with state-of-the-art on three benchmarks. Our method outperforms all the other methods.}
\label{comparison}
\end{figure}

\subsection{Comparison with the state-of-the-art\label{sec4.2}}
\textbf{Our method outperforms the state-of-the-art.} We compare our method with the state-of-the-art and the results are presented in Table \ref{mini_sota}, \ref{cifar_sota} and \ref{cub_sota} respectively. We compare the Top-1 accuracy in each session and report the performance dropping rate (PD) which measures the accuracy drop in the final session. w.r.t. the accuracy in the first session. We also show the performance curve in Figure \ref{comparison}. As shown in the tables, our method has the highest testing accuracy for each session in all three datasets. The performance dropping rate is relatively small. Especially, our ensemble model has an excellent performance on the base session, which demonstrate the power of the backbone in our training paradigm. In the last session, our method outperforms the SOTA method over $1.54\%$ on CIFAR100 dataset, $6.04\%$ improvement on \textit{mini}ImageNet dataset and $1.00\%$ on CUB200 dataset.

\begin{table}[]
\caption{Ablation study of our method on \textit{mini}ImageNet dataset.}
\vskip -0.2in
\label{ablation}
\begin{center}
\resizebox{\textwidth}{!}{

\begin{tabular}{cccccccccccll}
\toprule
\multirow{2}{*}{Ens.} & \multirow{2}{*}{SSL} & \multirow{2}{*}{DA} & \multicolumn{9}{c}{Acc. in each session(\%)} & \multirow{2}{*}{\textbf{PD} $\downarrow$} \\ \cline{4-12}
 &  &  & \multicolumn{1}{c}{1} & \multicolumn{1}{c}{2} & \multicolumn{1}{c}{3} & \multicolumn{1}{c}{4} & \multicolumn{1}{c}{5} & \multicolumn{1}{c}{6} & \multicolumn{1}{c}{7} & 8 & 9 &  \\
\midrule
 &  &  & 75.55 & 68.62 & 63.40 & 59.80 & 56.91 & 53.65 & 50.81 & 50.06 & 48.30 & 27.25 \\
\checkmark &  &  & 81.00 & 74.03 & 69.41 & 66.27 & 62.85 & 60.54 & 57.56 & 56.65 & 55.51 & 25.49 \\
\checkmark & \checkmark &  & 81.27 & 73.82 & 69.36 & 65.93 & 62.61 & 60.36 & 57.90 & 55.95 & 56.11 & 25.16 \\
\checkmark &  & \checkmark & \textbf{81.32} & \textbf{74.82} & 69.94 & \textbf{66.61} & 63.66 & 61.11 & 57.90 & 56.80 & 56.33 & 24.99 \\
\midrule
\checkmark & \checkmark & \checkmark & 81.28 & 74.29 & \textbf{70.07} & 66.51 & \textbf{63.80} & \textbf{61.40} & \textbf{57.99} & \textbf{57.04} & \textbf{56.53} & \textbf{24.75} \\
\bottomrule
\end{tabular}
}
\end{center}
\end{table}

\subsection{Ablation study}
To prove the effectiveness of our proposed method, we conduct several ablation experiments on \textit{mini}ImageNet to analyze the effectiveness of each components in our method. We keep the same basic implementation details as in Section \ref{sec4.1}. There are mainly three components in our method, including model ensembling(Ens.), self-supervised learning(SSL) and PatchMix(DA). We train the ResNet18 network with replay as our baseline. For incremental learning sessions, we fix the backbone and train the classifier where the weight vector of each new class is initialized by randomly choosing the data embeddings in the training set. We report the experimental results in Table \ref{ablation}. 

\begin{figure}
\centering
\includegraphics[width=0.4\textwidth]{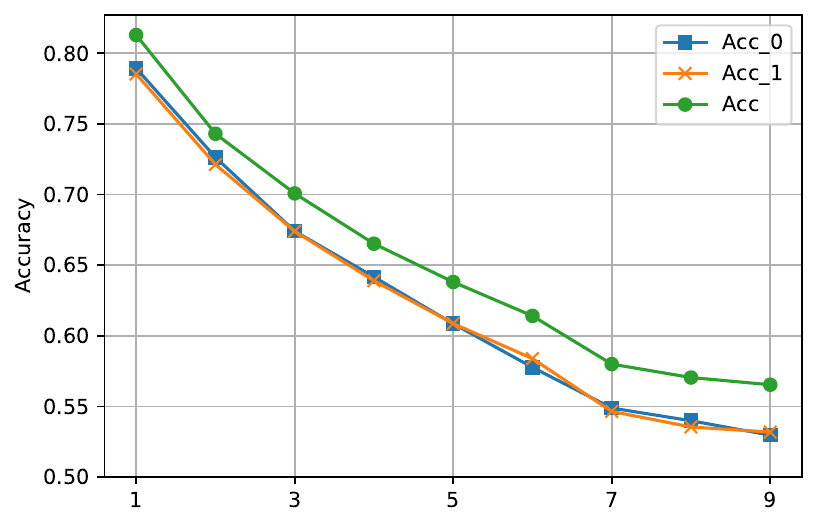}
\caption{Accuracy of each classifier and the whole model on \textit{mini}ImageNet.}
\label{ens_acc}
\end{figure}

As shown in Table \ref{ablation}, the ensembling training scheme most benefits the performance, which has an over $5\%$ improvement of accuracy on every session. We can also find that the dropping rate is reduced compared to the baseline. To see the detailed mechanism of model ensembling, we measure the accuracy curve of each classifier respectively in each session on \textit{mini}ImageNet. The result is shown in Figure \ref{ens_acc}. We can see from the figure that each separate classifier can outperform the state-of-the-art while combining the two classifiers can have a further promotion to the performance. We can infer that our ensemble model can have a calibration of the output and the bias can be avoided to a considerable extent. And from the deceleration of PD we can see that overfitting problem is mitigated. When combined with SSL, the model learns more affluent features thus a higher performance is obtained and the performance drop rate also decreases. When combined ensembling with PatchMix, better performance is obtained because DA can help increase the generalizability of the model. Finally, when we use all these three methods, the performance of the base model has a slight drop rate because the model has to carve out extra space to learn the SSL task. But the whole performance rises with the incremental learning sessions and the drop rate is also the lowest. This ablation study demonstrates that each part of our method is helpful for the FSCIL tasks.

\subsection{Further analysis}

\begin{figure}
	\begin{minipage}[t]{0.45\linewidth}
		\centering
		\includegraphics[width=0.9\textwidth]{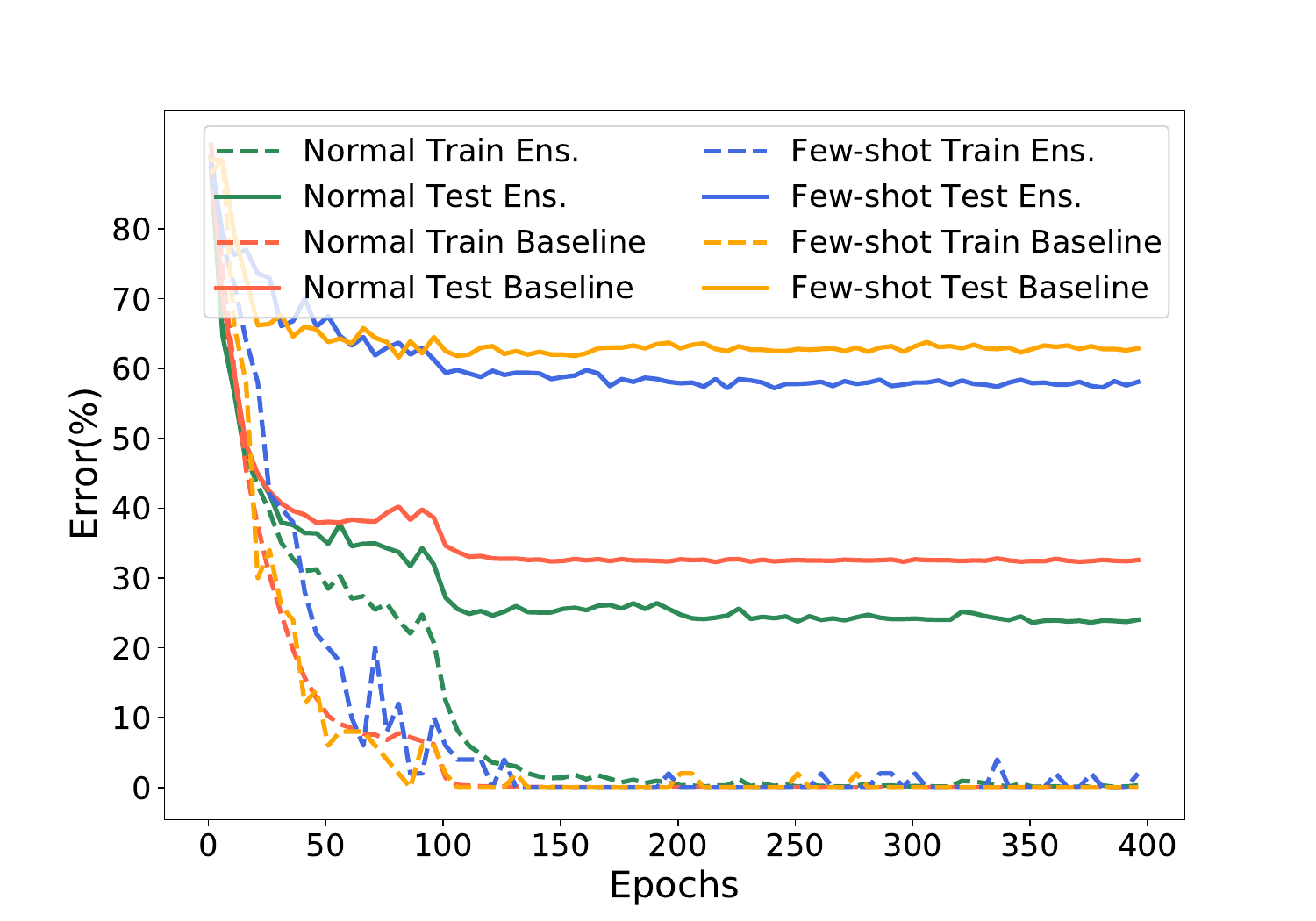}
		\caption{Overfitting problem in normal training and ensemble learning scenarios.}
		\label{overfitting}
	\end{minipage} ~~~~
	\begin{minipage}[t]{0.45\linewidth}
		\centering
		\includegraphics[width=0.8\textwidth]{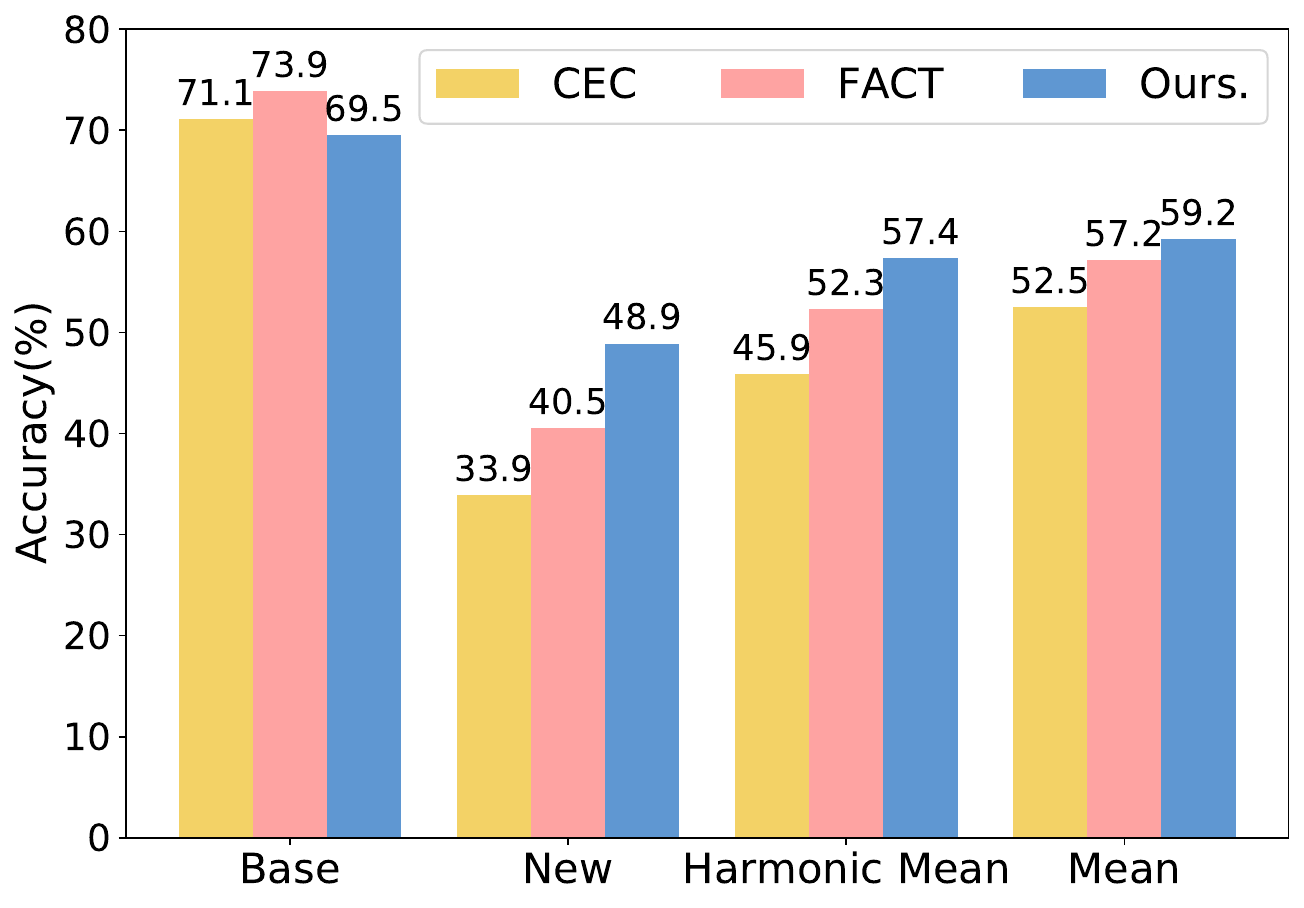}
		\caption{Accuracy of base and new classes on CUB200 dataset.}
		\label{new_acc}
	\end{minipage}
\end{figure}

\textbf{Overfitting problem is more severe in few-shot learning.}
We study the overfitting problem in normal learning and few-shot learning. We compare our method with standard ResNet18 model (baseline) on the two scenarios and plot the training accuracy and testing accuracy in the training process in Figure \ref{overfitting}. We train the base session on CIFAR100 as `Normal' while we randomly select 10-way-5-shot samples and train the models from scratch as `Few-shot'. As can be seen in the figure, overfitting problem is more severe in few-shot learning scenario. Although 10-classes classification problem is relatively easy, the gap between training and testing is extremely high. Compared to normal training, ensembling model requires more epochs to train but can alleviate overfitting to some extent.  

\textbf{Analysis of PatchMix module.} We investigate the role of PatchMix in FSCIL. In particular, we compare our spatial-aware sampling strategy with uniform sampling in different numbers of patches. We use a unified base model without DA as the pre-train model and the result is shown in Table \ref{mini_patches}. As we can see, spatial-aware PatchMix is more effective and can achieve good performance over all sessions compared to uniform sampling. The gap is more significant when more patches are selected. When the number of patches is between 3 and 5, the best performance is obtained. Thus we conclude our patchmix method is useful to avoid overfitting problem.

\textbf{Performance measure.} To better comprehend the learning result of FSCIL, we utilize more detailed performance measure, including the accuracy of base classes and new classes, harmonic mean accuracy and mean accuracy on CUB200 dataset. We compare our method with two SOTA methods, CEC and FACT. The results are reported in Figure \ref{new_acc}. As shown in the figure, although the base classes cannot exceed the two methods, the performance of new classes far surpasses other methods. This result shows our method can promote the learning of the few-shot sessions and alleviate overfitting.

\begin{table}[!t]
\caption{Study of different ways of samplings for FSCIL on \textit{mini}ImageNet.}
\vskip -0.1in
\label{mini_patches}
\begin{center}
\resizebox{0.8\textwidth}{!}{
\begin{tabular}{ccccccccccc}
\toprule
\multirow{2}{*}{\# patches} & \multirow{2}{*}{Sampling} & \multicolumn{9}{c}{Acc. in each session(\%)} \\ \cline{3-11} 
 &  & 1 & 2 & 3 & 4 & 5 & 6 & 7 & 8 & 9 \\
 \midrule
\multirow{2}{*}{$\mathcal U\left[1,3\right]$} & Uni. & 81.00 & 74.23 & 69.70 & 66.05 & 63.54 & 60.93 & 58.13 & 56.06 & \textbf{55.87} \\
 & Spa. & 81.00 & 74.20 & 69.29 & 66.33 & 63.48 & 60.68 & 57.89 & 56.37 & 55.67 \\
 \midrule
\multirow{2}{*}{$\mathcal U\left[3,5\right]$} & Uni. & 81.00 & 73.62 & 69.20 & 65.83 & 63.10 & 60.35 & 57.68 & 56.04 & 54.83 \\
 & Spa. & 81.00 & 74.28 & 69.87 & 66.35 & 63.68 & 61.32 & 57.84 & 56.98 & \textbf{56.23} \\
 \midrule
\multirow{2}{*}{$\mathcal U\left[5,7\right]$} & Uni. & 81.00 & 72.20 & 66.50 & 63.13 & 60.68 & 58.21 & 55.13 & 54.42 & 53.52 \\
 & Spa. & 81.00 & 74.00 & 69.26 & 66.07 & 63.52 & 60.67 & 57.77 & 56.43 & \textbf{55.31} \\
 \midrule
\multirow{2}{*}{$\mathcal U\left[7,9\right]$} & Uni. & 81.00 & 70.77 & 66.01 & 62.55 & 60.22 & 57.56 & 54.43 & 53.97 & 52.64 \\
 & Spa. & 81.00 & 74.05 & 69.03 & 65.93 & 63.44 & 61.08 & 57.93 & 56.37 & \textbf{55.81} \\
 \bottomrule
\end{tabular}
}
\end{center}
\end{table}

\section{Conclusion and future work\label{sec5}}

In this paper, we propose a novel learning paradigm for few-shot class-incremental learning problem. In our learning scheme, we leverage ensemble model with data augmentation and advanced self-supervised learning method to acquire the constructive base model, which benefits the learning of few-shot sessions and obviously reduces overfitting. Extensive experiments on the three benchmark datasets show that our method is effective which has a high performance on the base classes and effectively adapts to new classes and overcomes overfitting. 

\textbf{Limitations.} One limitation of our work is that we use one small model ensembling method as our backbone to reduce cost and pursue a fair comparison. However, large ensembling method could be used to improve the performance in reality such as using an ensemble of large-scale pre-trained models as in \cite{chowdhury2021few}. Another limitation is that we only test our method based on the standard benchmark setting. However, the problems maybe more complex and changing in real-world situations. Thus a more flexible setting will be considered in our future work.

\newpage
\bibliography{references} 
\newpage
\appendix


\section{Appendix}

In this part, we first introduce the self-supervised learning (SSL) method that we base on in Section \ref{ap1.1}. We give a more detailed description of experimental setup in Section \ref{ap1.2}. More experimental results are displayed in Section \ref{ap1.3}.

\subsection{Introduction to VICReg\label{ap1.1}}
We borrow ideas from the state-of-the-art SSL method VICReg \cite{bardes2021vicreg} to design our SSL training method adapted to ensemble model. Although we already show the training process of SSL, the concrete form of the SSL loss has not been introduced. We give a detailed description of VICReg here

\begin{figure}[htbp]
\centering
\includegraphics[trim=0cm 0.2cm 0cm 0cm, clip, width=157mm]{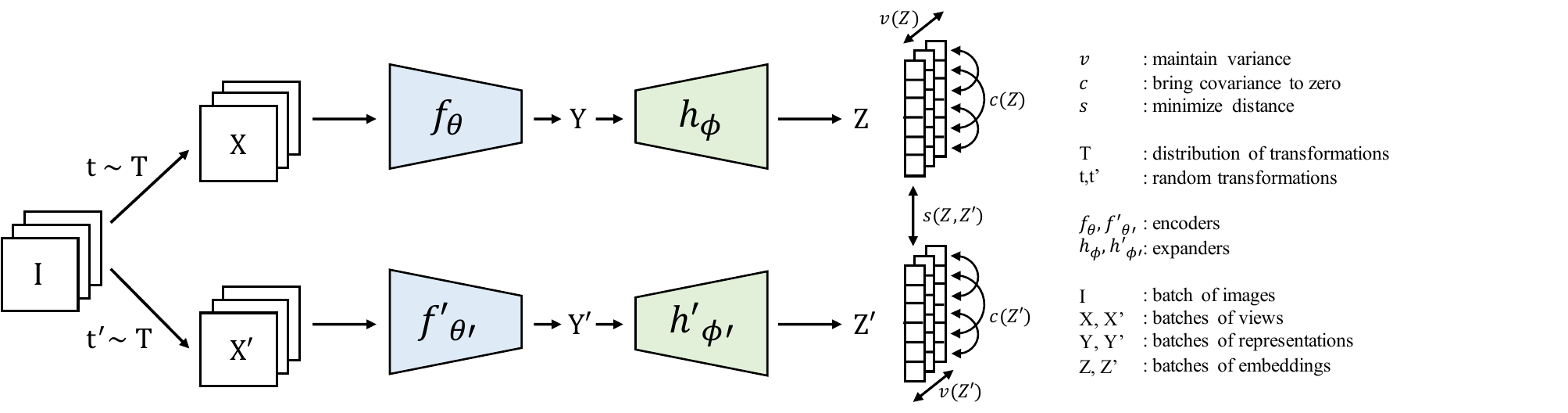}
\caption{The framework of VICReg.}
\label{vicreg}
\end{figure}

Figure \ref{vicreg} shows the framework of VICReg. Given an image, two transformations of $t$ and $t^{\prime}$ are sampled from a transformation distribution $\mathcal{T}$ and imposed to the image. Then the produced two images are encoded into the feature extractor and get two representations $y = f_{\theta}(t(x))$ and $y^{\prime} = f_{\theta}(t^{\prime}(x))$, which are then mapped onto the embeddings $z = h_{\phi}(y)$ and $z^{\prime} = h_{\phi}(y^{\prime})$ by the expander $h_{\phi}$. We denote $Z=\left[z_{1}, \ldots, z_{n}\right]$ and $Z^{\prime}=\left[z_{1}^{\prime}, \ldots, z_{n}^{\prime}\right]$. They are used for the computation of the loss. The loss has three terms: \textbf{variance}, \textbf{invariance} and \textbf{covariance}. The invariance is the mean square distance between the embedding vectors of two batches. The variance is a hinge loss to maintain the difference in the feature embeddings of one batch. The covariance decorrelates the embeddings in the batch and thus prevents model collapse by forcing the covariances over a batch between the embeddings torwards zero.

In specific, the loss function is as follows:
\begin{equation}
	\mathcal{L}_{\mathrm{SSL}}=\sum_{I\in \mathcal{D}}\sum_{t,t^{\prime}\sim \mathcal{T}}\lambda s\left(Z^{I}, Z^{\prime I}\right)+\mu\left[v(Z^{I})+v\left(Z^{\prime I}\right)\right]+\nu\left[c(Z^{I})+c\left(Z^{\prime I}\right)\right],
\end{equation}
where $\lambda,\mu$ and $\nu$ are hyper-parameters controlling the importance of each term.
 In this method, the parameters in the two network can be the same or different. In our implementation, we keep the same parameters. In another word, one batch of samples are doubled. Then the different transformations are imposed to the two batches of images. After that, the two batches are sent to the same network to compute the loss function. 
 
\subsection{More experimental details\label{ap1.2}}
\textbf{Training details.}
We train and evaluate the proposed model on a single tesla V100 GPU. We train the base model with batch size 128 on CIFAR100 and \textit{mini}ImageNet and 64 for CUB200 for 400 epochs. We use SGD with momentum 0.9, weight decay 0.0005 with Nesterov for optimization. For CIFAR100 and \textit{mini}ImageNet datasets, the learning rate starts from 0.0125 with multigamma step decay and warm-up in the base session. In incremental learning sessions, the model is trained for 20 epochs by a fixed backbone and the learning rate for fully-connected layer is still 0.0125. For CUB200, since the number of train samples in each class is about 30, a pre-trained model trained on ImageNet is often used. Thus we adopt a small learning rate 0.00125 with multigamma step decay in the base session. In incremental learning sessions, we use a different learning rate of 0.0005 on the backbone and 0.00125  on the classifier to control the change of the model.  

\textbf{Hyper-parameters.} (1) PatchMix: we adopt a $0.5$ probability to adopt PatchMix method. (2) SSL: For one batch of images, we adopt same transformation in the input. The transformation is randomly selected from rotation and flip. For the hyper-parameter $\gamma$ to control the importance of each loss, we set $\gamma$ to $0.2$ for CIFAR100 and \textit{mini}ImageNet datasets and $0.1$ for CUB200. The other weights are stay the same as in \cite{bardes2021vicreg}; As for the structure of the expander, we find that the model without expander can have a comparative performance. Thus we cancel the expander layer and calculate the SSL loss straight on the feature extracted by the backbone. 

\textbf{Comparison with replay-based method.} In this part, we discuss our exemplars replay strategy. The ideal situation in FSCIL is saving no samples and achieving unforgetting without storage cost. But in real-world applications, there are abundant datasets available, and the new coming data is rare in new coming classes. Thus we choose to keep a low memory bank that linearly increases with the number of classes. In our work, we save 5 samples for each base class and each new class. We also compare our method with existing replay-based methods. The SOTA method F2M \cite{shi2021overcoming} searches for a flat minimum and restricts the parameters change within a controllable range to reduce catastrophic forgetting. They saves 5 samples per new class to alleviate the forgetting of new classes. EKDIL \cite{dong2021few} saves 5 samples per class as prototypes for downstream tasks. In FSLL \cite{mazumder2021few}, we apply the results of the method that saves 5 samples for each class. All the results are shown in Table 1, 2, and 3 in our main submission. As shown in the tables, our method comparatively surpasses all the other methods.

\subsection{More experimental results\label{ap1.3}}
In this section, we conduct experiments to verify the effectiveness of our method and to show the inference cost. Firstly we make the whole training data the same as that in the paper \cite{tao2020few} and run our method on the three datasets 10 times by different random seeds to see the error bar. Secondly, we stay the same split of the classes in each session as in the paper \cite{tao2020few} and randomly sample 5-shot training data for each class from the original dataset. We conduct the experiments for 10 runs to see the stability of our method.

\textbf{Error bar of our method in different random seed.} We conduct our experiments for 10 runs in the incremental learning sessions on the three datasets. We keep the training data and the class orders the same as in \cite{tao2020few}. The means and the 95\% confidence are shown in Table \ref{seed}. We report the mean of these results in each session in our main submission and here we report the confidence interval for better verification. From the table we can see that the performance of our method is steady. The three datasets have similar performance. The variance is relatively small in the beginning and raises as the training sessions go on. We conclude it is because the randomness has an additive effect on the final result. The whole performance shows our method can steadily achieve the state-of-the-art.

\begin{table}[!h]
\caption{Classification accuracy on three datasets in different random seed with 95\% confidence intervals.}
\vskip -0.2in
\label{seed}
\begin{center}
\resizebox{\textwidth}{!}{
\begin{tabular}{lccccccccccc}
\toprule
\multirow{2}{*}{Dataset} & \multicolumn{11}{c}{Acc. in each session(\%).} \\ \cline{2-12} 
 \specialrule{0em}{1pt}{1pt} 
 & 1 & 2 & 3 & 4 & 5 & 6 & 7 & 8 & 9 & 10 & 11 \\
 \midrule
\multirow{2}{*}{CIFAR100} & \multirow{2}{*}{76.6} & 71.57 & 66.89 & 62.63 & 60.22 & 57.48 & 55.22 & 53.16 & 50.89 & - & - \\
 &  & $\pm$ 0.09 & $\pm$   0.13 & $\pm$   0.22 & $\pm$   0.24 & $\pm$   0.19 & $\pm$   0.23 & $\pm$   0.25 & $\pm$   0.26 & - & - \\
 \midrule
\multirow{2}{*}{\textit{mini}ImageNet} & \multirow{2}{*}{81.28} & 74.29 & 70.07 & 66.51 & 63.80 & 61.40 & 57.99 & 57.04 & 56.53 & - & - \\
 &  & $\pm$ 0.06 & $\pm$ 0.09 & $\pm$ 0.12 & $\pm$ 0.14 & $\pm$ 0.25 & $\pm$ 0.21 & $\pm$ 0.26 & $\pm$ 0.29 & - & - \\
 \midrule
\multirow{2}{*}{CUB200} & \multirow{2}{*}{80.83} & 76.77 & 71.98 & 68.27 & 67.78 & 66.27 & 63.42 & 61.98 & 60.05 & 58.65 & 57.09 \\
 &  & $\pm$ 0.13 & $\pm$ 0.21 & $\pm$ 0.24 & $\pm$ 0.32 & $\pm$ 0.35 & $\pm$ 0.37 & $\pm$ 0.33 & $\pm$ 0.29 & $\pm$ 0.36 & $\pm$ 0.31 \\
\bottomrule
\end{tabular}
}
\end{center}
\end{table}

\textbf{Error bar of our method in different training data.} We conduct experiments to see how different training data impact the final results in few-shot class-incremental problem. As shown in the table, these averages are different than that in Table \ref{data}. Different data can influence the performance obviously and variance is larger, especially in the earlier sessions. We consider it may be because the hedge effect of the randomness. 

\begin{table}[!h]
\caption{Classification accuracy on three datasets in different 5-way-5shot training data with 95\% confidence intervals.}
\vskip -0.2in
\label{data}
\begin{center}
\resizebox{\textwidth}{!}{
\begin{tabular}{lccccccccccc}
\toprule
\multirow{2}{*}{Dataset} & \multicolumn{11}{c}{Acc. in each session(\%).} \\ \cline{2-12} 
 \specialrule{0em}{1pt}{1pt} 
 & 1 & 2 & 3 & 4 & 5 & 6 & 7 & 8 & 9 & 10 & 11 \\
 \midrule
\multirow{2}{*}{CIFAR100} & \multirow{2}{*}{76.6} & 72.17 & 68.21 & 63.48 & 61.52 & 58.03 & 55.80 & 54.37 & 52.11 & - & - \\
 &  & $\pm$ 0.24 & $\pm$   0.22 & $\pm$   0.26 & $\pm$   0.21 & $\pm$   0.25 & $\pm$   0.26 & $\pm$   0.29 & $\pm$   0.31 & - & - \\
 \midrule
\multirow{2}{*}{\textit{mini}ImageNet} & \multirow{2}{*}{81.28} & 74.18 & 72.21 & 67.95 & 64.40 & 61.72 & 58.38 & 57.26 & 56.84 & - & - \\
 &  & $\pm$ 0.13 & $\pm$ 0.19 & $\pm$ 0.24 & $\pm$ 0.33 & $\pm$ 0.31 & $\pm$ 0.29 & $\pm$ 0.31 & $\pm$ 0.34 & - & - \\
 \midrule
\multirow{2}{*}{CUB200} & \multirow{2}{*}{80.83} & 77.21 & 73.43 & 69.78 & 68.26 & 67.02 & 64.86 & 63.12 & 61.79 & 60.13 & 59.42 \\
 &  & $\pm$ 0.23 & $\pm$ 0.25 & $\pm$ 0.23 & $\pm$ 0.31 & $\pm$ 0.37 & $\pm$ 0.39 & $\pm$ 0.34 & $\pm$ 0.35 & $\pm$ 0.31 & $\pm$ 0.34 \\
\bottomrule
\end{tabular}
}
\end{center}
\end{table}

\textbf{Inference cost compared to standard network.} 
To verify our lightweight ensemble model, we conduct experiments to compare our model inference time with standard ResNet18 network. We compute the inference time on the full test data in the three datasets. We run each experiment for 10 times and the average inference time is shown in Table \ref{t3}. As shown in the table, the computational gap between the two models is negligible.

\begin{table}[!h]
\caption{Inference time in three datasets}
\vskip -0.1in
\label{t3}
\begin{center}
\resizebox{0.5\textwidth}{!}{
\begin{tabular}{lccc}
\toprule
\multirow{2}{*}{} & \multicolumn{3}{c}{Inference time (\textit{ms})} \\
 \cline{2-4} 
 \specialrule{0em}{1pt}{1pt}
 & CIFAR100 & \textit{mini}ImageNet & CUB200 \\
\midrule
ResNet18 & 1.302 & 12.715 & 15.594 \\
\midrule
Our model & 1.319 & 12.840 & 15.670 \\
\bottomrule
\end{tabular}
}
\end{center}
\end{table}

\end{document}